\documentclass[letterpaper]{article} 
\usepackage{aaai25}  
\usepackage{times}  
\usepackage{helvet}  
\usepackage{courier}  
\usepackage[hyphens]{url}  
\usepackage{graphicx} 
\usepackage{natbib}  
\usepackage{caption} 
\usepackage{booktabs}
\usepackage{arabtex}
\usepackage{utf8}
\setcode{utf8}
\DeclareUnicodeCharacter{0627}
\title{EmoHopeSpeech: An Annotated Dataset of Emotions and Hope Speech in English and Arabic}
\author{
    Wajdi Zaghouani\textsuperscript{\rm 1},
    Md. Rafiul Biswas\textsuperscript{\rm 2}
}
\affiliations{
    \textsuperscript{\rm 1}Northwestern University in Qatar, Education City, Doha, Qatar\\
    \textsuperscript{\rm 2}Hamad Bin Khalifa University, Doha, Qatar\\
}
\renewenvironment{abstract}{
  \begin{center}
    \bf{Abstract}
  \end{center}
  \fontsize{9pt}{10pt}\selectfont
  \begin{quote}
  \setlength{\parindent}{0pt}
  \setlength{\parskip}{0pt}
}{
  \end{quote}
}
\begin{document}
\maketitle
\begin{abstract}
This research introduces a bilingual dataset comprising 27,456 entries for Arabic and 10,036 entries for English, annotated for emotions and hope speech, addressing the scarcity of multi-emotion (Emotion and hope). The dataset provides comprehensive annotations capturing emotion intensity, complexity, and causes, alongside detailed classifications and subcategories for hope speech. To ensure annotation reliability, Fleiss' Kappa was employed, revealing 0.75-0.85 agreement among annotators both for Arabic and English language. The evaluation metrics (F1-score: 0.69) obtained from the baseline model (i.e., machine learning model) validate that the data annotations are worthy for different analyses. This dataset offers a valuable resource for advancing natural language processing in underrepresented languages, fostering better cross-linguistic analysis of emotions and hope speech.
\end{abstract}
\section{Introduction}

In recent years, the proliferation of online content has emphasized the need for a deeper understanding of language and its impact on emotions and hope speech. The study of emotions in textual data has applications in various domains, including mental health monitoring \cite{dheeraj2021negative,fei2020deep}, customer sentiment analysis \cite{chintala2024emotion}, and personalized recommendations \cite{babanne2020emotion}. Similarly, hope speech, which refers to positive and uplifting expressions, plays a crucial role in promoting harmony, mitigating harmful narratives, and fostering inclusive communication \cite{balouchzahi2023polyhope}. Also, identifying and promoting hope speech could help counter the spread of toxic content and support digital well-being. 

Despite the significance of these areas, there remains a lack of comprehensive, multilingual datasets that cater to the analysis of both emotions and hope speech, particularly in underrepresented languages like Arabic. The inclusion of both English and Arabic in this dataset offers unique opportunities to explore cross-linguistic and cross-cultural comparisons in emotional expression and positivity. Arabic, as one of the most widely spoken languages globally, presents unique challenges and opportunities for emotion and hope speech detection due to its rich dialectal variations and complex morphological structure \cite{elnagar2021systematic}. 

Coupled with English, which is widely studied in NLP, this paper introduces EmoHopeSpeech, a bilingual annotated corpus that focuses on emotions and hope speech in English and Arabic. The dataset paves the way for dual language and cross-cultural studies, enriching the global research landscape \cite{gutierrez2016detecting}. The dataset is designed to capture the cultural and linguistic understandings of these two diverse languages. It includes labels for a variety of emotional categories as well as classifications for hope speech, providing researchers with a valuable resource for advancing natural language processing (NLP) models in these domains. We describe the dataset's construction, annotation process, and key characteristics, demonstrating its potential for advancing research in natural language processing, sentiment analysis, and cross-cultural communication. This work not only addresses the need for balanced datasets in low-resourced languages but also fosters the development of more inclusive and culturally aware AI systems.

Our dataset comprises carefully curated social media posts in both English and Arabic, annotated in a new direction of emotion analysis which consists of basic emotions, emotion intensity, complexity of emotion, and emotion causes. We labeled hope speech into binary label, and granular categories. The annotation process involved native speakers of both languages and followed a rigorous methodology to ensure high-quality labels. This dual-language approach enables comparative analysis of emotional expression patterns and hope speech characteristics across different cultural and linguistic contexts. This paper makes several key contributions to the field of natural language processing and affective computing:

\begin{itemize}
    \item The first large-scale dataset combining emotion and hope speech annotations in English and Arabic.
    \item A detailed analysis of annotation challenges and solutions specific to hope speech detection in multilingual contexts.
    \item Baseline models demonstrating the feasibility of joint emotion and hope speech detection.
    \item Insights into cross-cultural patterns of emotional expression and hope speech.
\end{itemize}

Our work lays the foundation for developing more NLP based content analysis systems that can identify not just negative content for moderation, but also positive, hope-inspiring content that could be amplified to create healthier online spaces. The remainder of this paper is organized as follows: Section 2 reviews related work in emotion detection and hope speech analysis. Section 3 describes the methodology for data collection, annotation, and preprocessing. Section 4 discusses the dataset structure and statistical insights. Section 5 presents potential applications and evaluation results. Finally, Section 6 concludes with future research directions.

\section{Related Works}
A good number of work has been done in the English language for Emotion analysis. Emotion analysis tasks can be classified in two ways. First is a publicly available dataset that is annotated with emotion labels. Second is the approach to developing machine learning and deep learning techniques to classify text into emotion labels. We will focus on the publicly available dataset for emotion recognition. While we discuss datasets, we need to comply with different categories: textual, voice, image, memes, and multimodal (image+text). \citep{yang2023emoset} collected 3 million images, out of which humans annotated 118,102 images and became the largest emotion dataset. Table \ref{tab:datasets} shows the recent work on emotion dataset.

\begin{table}[h!]
\scriptsize 
\centering
\begin{tabular}{|p{1cm}|l|p{2cm}|l|l|}
\hline
\textbf{Dataset} & \textbf{Modality} & \textbf{Annotation Label} & \textbf{Size} & \textbf{Language} \\ \hline
EmoSet (\cite{yang2023emoset}) & Image & Amusement, Anger, Awe, Contentment, Disgust, Excitement, Fear, Sadness & 118,102 & English \\ \hline
Semval-2019 (\cite{chatterjee-etal-2019-semeval}) & Text & Angry, Happy, Sad & 38,424 & English \\ \hline
ISEAR (\cite{scherer1994evidence}) & Text & Joy, Sadness, Fear, Anger, Guilt, Disgust, and Shame & 7,665 & English \\ \hline
CARER (\cite{saravia-etal-2018-carer}) & Text & Anger, Anticipation, Disgust, Fear, Joy, Sadness, Surprise, and Trust & 2,000 & English \\ \hline
Arabic (\cite{teahan2019new}) & Text & Anger, Disgust, Fear, Happiness, Sadness and Surpris & 1221 & Arabic \\ \hline
\end{tabular}
\caption{Summary of datasets with their modalities, annotation labels, sizes, and languages.}
\label{tab:datasets}
\end{table}

\citealp{chakravarthi2020hopeedi} developed a multilingual hope speech dataset of 6176 entries in English, Tamil, and Malayalam languages. \citealt{balouchzahi2023polyhope} collected English tweets and created a hope speech dataset. \citeauthor{divakaran2024hope} build a hope speech dataset from Reddit posts comprising of two languages Spanish and English. 
However, according to the best knowledge of the author, no speech work has been done in Arabic. We aimed to create a large scale  hope speech dataset for hope speech. Moreover, the integration of emotion annotation with hope speech meaningful insight into public sentiment. 
\section{Data Collection and Annotation}
\subsection{Data Collection}
We collected publicly available emotion datasets in Arabic and English that were labeled with basic emotion labels (happiness, sad, love, joy, amused, disgust, fear, empathy, and confidence). We collected publicly available arabic dataset from three different sources i) Arabic Poetry emotions\cite{shahriar2023classification}, ii) Emotional Tone\cite{al2018emotional}, iii) Multilabel Hate speech dataset\cite{zaghouani2024so}. The Arabic Poetry emotions dataset has 9452 rows, the Emotional Tone dataset has 10065 rows, and the Hate speech dataset has 30000 rows. In total, 49517 Arabic data were labeled with emotions. We kept rows in which text data had more than 5 words and less than 80 words. After filtering out, 27456 rows remain. To keep track of the original sources of data, we have added a column for sources that keeps the URL of the original data. 

For English dataset, we collected from Emotion analysis based on text analysis\cite{anjali2024emotion}. It is a large-volume dataset with 393822 rows and annotated basic emotion labels. After filtering with min and max words, it left with 10036 rows. Among them, we could annotate 4036 rows and include them in this study for further analysis into hope speech.

\subsection{Data Annotation Process}
The dataset was initially labeled with the basic emotion label. We employed 10 native Arabic speakers to annotate the Arabic dataset into different types of emotion labels: Emotion Intensity, Complexity of Emotion, and Emotion Cause and hope speech labels. The native Arabic speakers are male and female from different Arab countries such as Qatar, Tunisia, Jordan, and Egypt maintaining the diversity. It also ensures the dialectal variation of data annotation. A manager manages the data annotation process. There were several training sessions to train the annotators. Annotators meet frequently to discuss their progress and resolve conflicts through discussions. Similarly, for English data, five annotators involve in this data annotation task. Guidelines was provided with proper example both in Arabic and English. Guidelines was revised several times to make it clear for understanding. 

\textbf{Emotion Intensity:} Table \ref{tab:emotion_intensity} provides a structured framework for categorizing the intensity of emotions expressed in both English and Arabic texts. It has four categories: Low, Medium, High, Not Applicable. Annotator assess the intensity or strength of the emotion expressed in the text and select the label that best matches the emotional intensity. 

\begin{table}[ht]
\scriptsize 
\centering
\begin{tabular}{|p{1cm}|p{1.5cm}|p{2cm}|p{2cm}|}
\hline
\textbf{Label} & \textbf{Guidelines} & \textbf{English Example} & \textbf{Arabic Example} \\ \hline
Low & The emotion is present but not strongly expressed & "I'm a bit worried about the test tomorrow, but it's not a big deal." & \begin{arabtext}أنا قلق قليلاً بشأن الامتحان غدًا، لكن الأمر ليس مهمًا\end{arabtext} \\ \hline
Medium & The emotion is clear and noticeable but not overwhelming & "I'm worried about the test tomorrow; it's making me feel uneasy." & \begin{arabtext}
    أنا قلق بشأن الامتحان غدًا؛ يجعلني أشعر بعدم الارتياح.
\end{arabtext} \\ \hline
High & The emotion is intense and strongly expressed & "I'm extremely worried about the test tomorrow! I can't stop thinking about it, and it's driving me crazy!" & \begin{arabtext}
    أنا قلق للغاية بشأن الامتحان غدًا! لا أستطيع التوقف عن التفكير فيه، إنه يجننني!
\end{arabtext} \\ \hline
Not Applicable (N/A) & No emotion is present or intensity cannot be determined & "The test is scheduled for tomorrow." & \begin{arabtext}
الامتحان مجدول ليوم غد
\end{arabtext}\\ \hline
\end{tabular}
\caption{Emotion Intensity Table}
\label{tab:emotion_intensity}
\end{table}

\textbf{Emotion Complexity}: The Table \ref{tab:emotion_complexity} framework demonstrates the complexity of emotion. There are four categories of emotion complexity: simple, medium, complex, N/A (Not Applicable). The task is to evaluate whether the emotion in the text is simple or complex. 

\begin{table}[ht]
\scriptsize 
\centering
\begin{tabular}{|p{1cm}|p{2cm}|p{1.5cm}|p{1.5cm}|}
\hline
\textbf{Label} & \textbf{Guidelines} & \textbf{English Example} & \textbf{Arabic Example} \\ \hline
Simple & The text clearly expresses a single, straightforward emotion. & I'm so happy today! & \begin{arabtext}أنا سعيد جدًا اليوم \end{arabtext}\\ \hline
Medium & The emotion is more nuanced than simple, but it's still relatively clear and not as complex as juggling multiple distinct emotions. & I’m feeling a bit frustrated, but it’s not too bad. & \begin{arabtext}
    أشعر بالإحباط قليلاً، لكن الأمر ليس سيئًا للغاية
\end{arabtext}\\ \hline
Complex & The text expresses multiple emotions or a nuanced emotional state. & I’m excited about the new job, but also nervous about leaving my old one. & \begin{arabtext}أنا متحمس للعمل الجديد، لكنني أشعر بالتوتر أيضًا بشأن ترك عملي القديم \end{arabtext}\\ \hline
Not Applicable (N/A) & No clear emotion or complexity cannot be determined. & The meeting is scheduled for 3 PM. & \begin{arabtext}   
الاجتماع مجدول في الساعة الثالثة مساءً \end{arabtext} \\ \hline
\end{tabular}
\caption{Complexity of Emotion Table}
\label{tab:emotion_complexity}
\end{table}

\textbf{Emotion Cause:} The table \ref{tab:emotion_cause} provides a framework for annotators to classify text based on what is causing the expressed emotion. It is used in sentiment analysis or emotion recognition tasks where understanding the cause is important.
\begin{table}[ht]
\scriptsize 
\centering
\begin{tabular}{|p{1cm}|p{2cm}|p{2cm}|p{2cm}|}
\hline
\textbf{Label} & \textbf{Guidelines} & \textbf{English Example} & \textbf{Arabic Example} \\ \hline
External Event & The emotion is caused by an event or situation outside the individual. & I felt scared when I heard the loud thunderstorm outside. & \begin{arabtext}شعرت بالخوف عندما سمعت العاصفة الرعدية القوية في الخارج\end{arabtext} \\ \hline
Internal Reflection & The emotion arises from internal thoughts or self-reflection. & I feel so guilty for the mistakes I've made in the past. & \begin{arabtext}
    أشعر بالذنب الشديد بسبب الأخطاء التي ارتكبتها في الماضي
\end{arabtext}  \\ \hline
Relationship & The emotion is linked to interactions or relationships with others. & I was heartbroken when my friend stopped talking to me. & \begin{arabtext}
    شعرت بالحزن الشديد عندما توقف صديقي عن الحديث معي
\end{arabtext}\\ \hline
Achievement Or Failure & The emotion is related to personal success or failure. & I was overjoyed when I got the promotion at work. & \begin{arabtext}
    كنت سعيدًا جدًا عندما حصلت على الترقية في العمل
\end{arabtext}\\ \hline
Unclear & The cause of the emotion is not clear. & I’ve been feeling really down lately, but I don’t know why. & \begin{arabtext}
    أشعر بالاكتئاب في الفترة الأخيرة، لكن لا أعرف السبب 
\end{arabtext} \\ \hline
Not Applicable (N/A) & No clear emotion or cause is stated. & The project is due next week. & \begin{arabtext}
    المشروع يجب أن يُسلم الأسبوع القادم
\end{arabtext}\\ \hline
\end{tabular}
\caption{Emotion Cause Table}
\label{tab:emotion_cause}
\end{table}

\textbf{Hope Speech:} Hope Speech in Table \ref{tab:speech_categories} categorizes texts that express hope, positivity, encouragement, or a constructive outlook as "Hope Speech," differentiating them from "Not Hope Speech," which lacks these elements. The table also includes "Counter Speech," which challenges negativity or hate speech constructively, and "Neutral" texts, which are factual or devoid of emotional tone. Additionally, it defines "Hate Speech or Negativity" to capture texts that are offensive, harmful, or convey negative sentiments. The table provides examples in both English and Arabic to ensure clarity and applicability across languages, making it a valuable tool for annotating and analyzing multilingual datasets.

\begin{table}[ht]
\scriptsize
\centering
\begin{tabular}{|p{1cm}|p{2cm}|p{2cm}|p{2cm}|}
\hline
\textbf{Label} & \textbf{Definition} & \textbf{Arabic Example} & \textbf{English Example} \\ \hline
Hope Speech & Express hope, positivity, encouragement, or a constructive outlook. & \begin{arabtext} ننجح في النهاية بغض النظر عن الصعوبات\end{arabtext} & We will succeed in the end despite the difficulties. \\ \hline
Not Hope Speech &  Does not contain any elements of hope or positivity. & \begin{arabtext}
    لا توجد تطورات جديدة
\end{arabtext}& There are no new developments. \\ \hline
Counter Speech &   Hate speech or negativity with the intention of refuting or challenging it. & \begin{arabtext}
    يجب علينا أن نرفض كل أشكال الكراهية
\end{arabtext} & We must reject all forms of hatred. \\ \hline
Neutral & Neutral or factual in nature, without conveying hope, positivity, or countering negative speech. & \begin{arabtext}
    المؤتمر سيعقد غدًا في الساعة التاسعة صباحًا
\end{arabtext} & The conference will be held tomorrow at 9 AM. \\ \hline
Hate Speech or Negativity & Offensive or harmful text, often targeting individuals or groups such as race, religion, ethnicity.  & \begin{arabtext}أمثالك هم السبب في كل المشاكل. يجب أن تشعر بالخجل \end{arabtext} & People like you are the cause of all the problems. You should be ashamed. \\ \hline
\end{tabular}
\caption{Categories of Hope Speech}
\label{tab:speech_categories}
\end{table}

\textbf{Hope Speech Subcategories:} The table \ref{tab:hope_speech_themes} categorizes hope speech into four distinct themes, each with its unique focus and purpose. Inspirational/Motivational emphasizes uplifting and motivating individuals, encouraging perseverance in the face of challenges, as seen in messages about achieving the impossible with determination and persistence. Solidarity/Peace highlights unity, community support, collective action, and the promotion of peace. Lastly, Spiritual/Empowerment includes speech inspired by religious or spiritual beliefs. 
\begin{table}[ht!]
\scriptsize
\centering
\begin{tabular}{|p{1cm}|p{2cm}|p{2cm}|p{2cm}|}
\hline
\textbf{Label} & \textbf{Definition} & \textbf{Arabic Example} & \textbf{English Example} \\ \hline
Inspiration & Inspire and motivate individuals and encourage them to keep going despite challenges. & \begin{arabtext}
    العزيمة والإصرار، يمكننا تحقيق المستحيل
\end{arabtext} & With determination and persistence, we can achieve the impossible. \\ \hline
Solidarity & Emphasizes community support, collective action, and promotion of peace and reconciliation & \begin{arabtext}
    بالتعاون بيننا، سنتمكن من تجاوز كل العقبات
\end{arabtext} & Through our cooperation, we will overcome all obstacles. \\ \hline
Resilience & Focuses on the resilience of individuals or communities and the long-term vision for a better future & \begin{arabtext}
    نحن نبني مستقبلًا مشرقًا لأجيالنا القادمة
\end{arabtext} & We are building a bright future for the next generations. \\ \hline
Spiritual & Religious or spiritual beliefs to instill hope. & \begin{arabtext}
    لكل واحد منا القدرة على تغيير واقعه للأفضل
\end{arabtext} & Each one of us has the power to change our reality for the better. \\ \hline
\end{tabular}
\caption{Subcategories of Hope Speech}
\label{tab:hope_speech_themes}
\end{table}

\section{Statistical Analysis}
\textbf{Emotion counts:} Table \ref{tab:emotion_counts} provides a summary of emotion counts categorized by language (Arabic and English) along with their respective proportions in the dataset. Each row represents a specific emotion label (e.g., sadness, joy, anger), with the count (N) of occurrences and the corresponding percentage (\%) of that label in the dataset for each language. For instance, "Sad" appears 2929 times (10.6\%) in Arabic and 84 times (2\%) in English, while "Disgust" is the most frequent label in Arabic with 6387 occurrences (23\%) and 75 occurrences (0.8\%) in English. Conversely, the "Neutral/None" category is dominant in English with 3261 occurrences (80\%), reflecting the variability in label distribution across languages. This analysis highlights the differences in emotion representation between Arabic and English datasets.

\begin{table}[ht]
\scriptsize
\centering
\begin{tabular}{|l|c|c|}
\hline
\textbf{Label} & \textbf{Arabic (N) (\%)} & \textbf{English (N) (\%)} \\ \hline
Sad            & 2929 (0.106)            & 84 (0.02)                 \\ 
Love           & 2855 (0.104)            & 178 (0.04)                \\ 
Joy            & 2229 (0.08)             & 184 (0.08)                \\ 
Surprise       & 1775 (0.06)             & 36 (0.008)                \\ 
Disgust        & 6387 (0.23)             & 75 (0.008)                \\ 
Fear           & 1743 (0.06)             & 23 (0.005)                \\ 
Empathetic     & 1534 (0.05)             & 77 (0.01)                 \\ 
Anger          & 1735 (0.06)             & 50 (0.01)                 \\ 
Neutral/None   & 3620 (0.13)             & 3261 (0.8)                \\ 
Optimism       & 1121 (0.04)             & 38 (0.009)                \\ 
Trust          & 1477 (0.05)             & 25 (0.006)                \\ \hline
\end{tabular}
\caption{Emotion counts by language.}
\label{tab:emotion_counts}
\end{table}

Table \ref{tab:emotion_summary} presents a comparative analysis of the Complexity of Emotion, Emotion Intensity, and Emotion Cause across Arabic and English datasets. It highlights the variation in the frequency and proportion of different categories between the two languages. For instance, Arabic data shows higher counts for categories such as "Medium Complexity" (14,325, 52\%) and "Internal Reflection" (10,857, 39\%) compared to English, where the respective counts are 1,137 (29\%) and 1,276 (32\%). In contrast, English exhibits a higher percentage for "Simple Complexity" (49\%) and "Low Intensity" (24\%) than Arabic (24\% and 12\%, respectively). These differences suggest language-based cultural and linguistic nuances in expressing emotions, with Arabic data skewing towards more complex and reflective emotional categories. The values bear significance as they inform the design of emotion-based models and emphasize the need for language-specific approaches in NLP tasks to ensure nuanced understanding and accurate representations.

\begin{table}[ht]
\scriptsize
\centering
\begin{tabular}{|l|c|c|}
\hline
\textbf{Category} & \textbf{Arabic (N) (\%)} & \textbf{English (N) (\%)} \\ \hline
\multicolumn{3}{|c|}{\textbf{Complexity of Emotion}} \\ 
Medium            & 14325 (0.52)            & 1137 (0.29)               \\ 
Complex           & 6575 (0.239)            & 867 (0.21)                \\ 
Simple            & 6546 (0.238)            & 1942 (0.49)               \\ \hline
\multicolumn{3}{|c|}{\textbf{Emotion Intensity}} \\ 
Medium            & 13611 (0.495)           & 1862 (0.46)               \\ 
High              & 10517 (0.383)           & 1155 (0.29)               \\ 
Low               & 3320 (0.12)             & 963 (0.24)                \\ \hline
\multicolumn{3}{|c|}{\textbf{Emotion Cause}} \\ 
Internal Reflection          & 10857 (0.39)            & 1276 (0.32)               \\ 
Relationship                 & 11124 (0.40)            & 785 (0.19)                \\ 
External Event               & 3377 (0.12)             & 968 (0.24)                \\ 
Achievement/Failure          & 1099 (0.04)             & 407 (0.10)                \\ 
Unclear                     & 997 (0.03)              & 551 (0.14)                \\ \hline
\end{tabular}
\caption{Summary of Complexity of Emotion, Emotion Intensity, and Emotion Cause by Language}
\label{tab:emotion_summary}
\end{table}

\textbf{Hope Speech:} Table \ref{tab:hope_speech} provides a detailed comparison of Hope Speech Categories and Hope Speech Subcategories across Arabic and English datasets, highlighting both the distribution and variation in proportions.

In the Hope Speech Categories, the "Neutral" category dominates in both Arabic (9,278, 34\%) and English (1,894, 46\%), indicating a significant presence of neutral expressions in the data. The "Hope Speech" category, which reflects positive and uplifting content, is more prevalent in Arabic (3,375, 12\%) compared to English (574, 14\%). Conversely, the "Negative/Hate Speech" and "Counter Speech" categories show lower proportions, with Arabic exhibiting a higher count (4,578, 16\% and 1,178, 4\%, respectively) than English (212, 5\% and 38, 0.009\%).

In the Hope Speech Subcategories, "Inspirational/Motivational" dominates in both languages, accounting for 39\% in Arabic (1,270) and 48\% in English (279). Subcategories like "Spiritual Empowerment" and "Solidarity/Peace" show similar trends, although Arabic has higher counts overall. Notably, "Resilience/Visionary" is represented proportionally higher in English (20\%) compared to Arabic (16\%).

\begin{table}[ht]
\scriptsize
\centering
\begin{tabular}{|p{3cm}|p{2cm}|p{2cm}|}
\hline
\textbf{Category} & \textbf{Arabic (N) (\%)} & \textbf{English (N) (\%)} \\ \hline
\multicolumn{3}{|c|}{\textbf{Hope Speech Categories}} \\ 
Neutral                   & 9278 (0.34)            & 1894 (0.46)               \\ 
Hope Speech               & 3375 (0.12)            & 574 (0.14)                \\ 
Not Hope Speech           & 9043 (0.33)            & 1318 (0.32)               \\ 
Hate Speech      & 4578 (0.16)            & 212 (0.05)                \\ 
Counter Speech            & 1178 (0.04)            & 38 (0.009)                \\ \hline
\multicolumn{3}{|c|}{\textbf{Hope Speech Subcategories}} \\ 
Inspirational & 1270 (0.39)            & 279 (0.48)                \\ 
Spiritual       & 864 (0.26)             & 94 (0.16)                 \\ 
Solidarity           & 543 (0.17)             & 83 (0.14)                 \\ 
Resilience      & 539 (0.16)             & 119 (0.2)                 \\ \hline
\end{tabular}
\caption{Summary of Hope Speech Categories and Subcategories by Language}
\label{tab:hope_speech}
\end{table}

\section{Data Analysis}
\subsection{Corelation between Emotion and Hope Speech:} Table \ref{tab:chi_square_results} summarizes the results of chi-square tests that analyze the connections between different emotional characteristics—such as Emotion Intensity(EI), Complexity of Emotion (CM), Emotion Cause(EC) and categories of hope speech (HS),
Hope Speech Subcategories(HSC) examined in both Arabic and English texts. It includes the chi-square statistics and corresponding p-values for each comparison. Notably, all relationships show statistically significant results (p < 0.05), with many having p-values of 0, reflecting very strong significance. The chi-square values vary across the comparisons, indicating different levels of association. Among the features analyzed, the relationship between emotion labels and hope speech categories stands out with the highest chi-square values for Arabic (1502.6) and English (160.5), highlighting the strongest association observed.

\begin{table}[ht]
\centering
\scriptsize 
\begin{tabular}{lcc}
\toprule
\textbf{Relationship} & \textbf{Arabic ($\chi^2$, P-Value)} & \textbf{English ($\chi^2$, P-Value)} \\
\midrule
EI vs. HC & 324.3, 0 & 194.7, 0 \\
EI vs. HSC & 72.9, 0 & 16, 0.004 \\
CM vs. HS & 354.3, 0 & 56.3, 0 \\
CM vs. HSC & 40.7, 0 & 43.1, 0 \\
EC vs. HC & 810.3, 0 & 245.6, 0 \\
EC vs. HSC & 121.3, 0 & 88.4, 0 \\
\bottomrule
\end{tabular}%
\caption{Relationships between Emotion Features and Hope Speech}
\label{tab:chi_square_results}
\end{table}

\subsection{Dataset Evaluation}
We analyzed the dataset annotations using both traditional machine learning methods, such as logistic regression and multinomial Naive Bayes, alongside modern transformer-based models like BERT. The preprocessing workflow included steps such as tokenization, cleaning the text, and transforming it into formats ready for modeling. To improve data quality, we eliminated stopwords from both Arabic and English datasets. For the Arabic dataset, we relied on the stopword list provided by \cite{alrefaie2019arabic}, and for the English dataset, we used tools from the NLTK library.

For model implementation, we fine-tuned AraBERT \cite{antoun2020arabert}, a pre-trained transformer model designed specifically for Arabic natural language processing tasks, and BERT-based-uncased \cite{DBLP:journals/corr/abs-1810-04805} for the English dataset. Both models were trained on the Hope Speech dataset to classify categories, leveraging their language-specific optimizations and contextual embeddings to boost prediction accuracy. Table \ref{tab:hope_speech_results} shows the model evaluation on our annotated dataset. Our annotated dataset served as a key resource for evaluating how well models could predict Hope Speech Categories. Performance metrics like precision, recall, F1-score, and accuracy provided a well-rounded assessment of how effectively the models aligned with the annotated labels. This evaluation not only validated the robustness of the annotation process but also underscored the role of advanced models in enhancing predictive accuracy for hope speech detection tasks.

\begin{table}[ht]
\scriptsize
\centering
\begin{tabular}{|p{1cm}|p{1cm}|p{1cm}|p{1cm}|p{1cm}|}
\hline
\multicolumn{5}{|c|}{\textbf{Hope Speech Prediction - Arabic Data}} \\ \hline
\textbf{Model}           & \textbf{Precision} & \textbf{Recall} & \textbf{F1-Score} & \textbf{Accuracy} \\ 
LR     & 0.43               & 0.45            & 0.42              & 0.45              \\ 
AraBERT                  & 0.49               & 0.49            & 0.47              & 0.49              \\ 
NB            & 0.44               & 0.45            & 0.41              & 0.45              \\ \hline
\multicolumn{5}{|c|}{\textbf{Hope Speech Prediction - English Data}} \\ \hline
LR     & 0.40               & 0.44            & 0.39              & 0.44              \\ 
AraBERT                  & 0.44               & 0.49            & 0.45              & 0.49              \\ 
NB            & 0.33               & 0.49            & 0.35              & 0.49              \\ \hline
\end{tabular}
\caption{Performance Metrics for Hope Speech Prediction - Arabic and English Data}
\label{tab:hope_speech_results}
\end{table}

\section{Error Analysis}
We provided a sample of 200 instances to all annotators and instructed them to annotate the data. Since multiple annotators were involved and the label are categorical, we calculated Fleiss' Kappa scores to measure the level of agreement. Fleiss' Kappa evaluates how consistently a group of annotators agrees on categorical data, accounting for agreement expected by chance. It ranges from -1 (complete disagreement) to 1 (perfect agreement), with 0 indicating agreement purely by chance. Table \ref{tab:fleiss_kappa} shows that Arabic data are generally higher than those of English, reflecting better consistency among annotators for Arabic annotations. Ranging from fair to moderate agreement depending on the category, the scores highlight differences in how annotators interpreted and labeled the data. Labels such as Emotion Cause (Ar-0.43, En-0.39) and Hope Speech Subcategories (Ar-0.36, En-).31) have lower kappa scores, indicating difficulties in interpreting and agreeing on these more subjective features.

\begin{table}[ht!]
\centering
\renewcommand{\arraystretch}{1.5}
\begin{tabular}{lcc}
\hline
\textbf{Label} & \textbf{Arabic} & \textbf{English} \\ \hline
Emotion Intensity         & 0.52 & 0.44 \\ 
Complexity of Emotion     & 0.50 & 0.47 \\ 
Emotion Cause             & 0.43 & 0.39 \\ 
Hope Speech Categories    & 0.56 & 0.48 \\ 
Hope Speech Subcategories & 0.36 & 0.31 \\ \hline
\end{tabular}
\caption{Inter-Annotator Agreement (IAA) - Fleiss' Kappa for Arabic and English Data}
\label{tab:fleiss_kappa}
\end{table}

In the following, we discuss the potential reasons for these variations and their implications for the annotation process.

\begin{itemize}
    \item Subjectivity in Labeling: Emotions and hope speech often involve subjective judgment, leading to varied interpretations. This is especially true for nuanced categories like intensity and complexity.

    \item Linguistic and Cultural Variations: Arabic and English texts may convey emotions and hope differently due to linguistic structures and cultural contexts. Annotators with different cultural backgrounds may interpret the same content in diverse ways.

    \item Ambiguity in Guidelines: If the annotation guidelines lacked sufficient clarity or examples, annotators may have interpreted criteria differently, resulting in lower agreement.

    \item Experience and Expertise of Annotators: Differences in annotators' familiarity with the subject matter, cultural context, or the guidelines could have contributed to inconsistent labeling.

    \item Complexity of Texts: Texts with subtle or indirect expressions of emotion, as well as those involving sarcasm, idiomatic expressions, or mixed sentiments, likely increased disagreement among annotators.
    
\end{itemize}

\section{Conclusion and Future Work}
In this paper, we integrated multiple datasets annotated with the basic emotion label. After cleaning the data, we annotated the dataset to include different variations of emotions, such as emotion intensity, emotion cause, and emotion complexity. Moreover, we annotated it to the hope speech and its subcategory. The annotation of emotion and hope speech demonstrates a significant relationship between hope speech and emotions.  Analyzing inter-annotator agreement signifies challenges and potential in predicting hope speech categories and emotional features in Arabic and English texts. The moderate Fleiss' Kappa scores point to the need for clearer annotation guidelines and more comprehensive training to improve agreement on complex and nuanced labels. Moving forward, efforts should prioritize expanding the dataset, incorporating diverse linguistic features, and exploring advanced pre-trained models and ensemble methods tailored for cross-linguistic tasks. Additionally, using data augmentation techniques and refining preprocessing approaches can further enhance model accuracy and annotation consistency, leading to better insights and tools for understanding emotional and hope-related speech across multiple languages.

\bibliography{ref}
\section{Ethics and Impact}
This study addresses critical ethical considerations and the potential societal impact of developing annotated datasets and predictive models for emotion and hope speech classification in Arabic and English. The dataset creation and model development were guided by a commitment to fairness, inclusivity, and cultural sensitivity. Original sources of data were included in the dataset to cite the prior work contribution. It does not contain any information related individual identity and so it exempt from Institutional Review Board
(IRB) approvals. Special care was taken to ensure that the annotation process reflected diverse perspectives and avoided bias in categorizing hope speech and emotions, particularly in sensitive contexts such as hate speech and counter-speech. Annotators were provided clear guidelines to minimize subjective inconsistencies and mitigate potential harms stemming from misrepresentation. 

\section{Dataset Release}
We have made dataset publicly available to the \url{10.5281/zenodo.14669301} and code repository
\url{github.com/rafiulbiswas/hopespeech} under the creative common license legalcode \url{creativecommons.org/licenses/by/4.0/legalcode}. We have shared two csv file containing Arabic and English data separately.

\end{document}